\documentclass{article}

\usepackage{PRIMEarxiv}

\usepackage[utf8]{inputenc} % allow utf-8 
\usepackage{hyperref}
\usepackage{CJKutf8}
\usepackage[T1]{fontenc}    % use 8-bit T1 fonts
\usepackage{hyperref}       % hyperlinks
\usepackage{url}            % simple URL typesetting
\usepackage{booktabs}       % professional-quality tables
\usepackage{amsfonts}       % blackboard math symbols
\usepackage{nicefrac}       % compact symbols for 1/2, etc.
\usepackage{microtype}      % microtypography
\usepackage{lipsum}
\usepackage{fancyhdr}       % header
\usepackage{graphicx}       % graphics
\graphicspath{{media/}}     % organize your images and other figures under media/ folder
\usepackage{subcaption}
\usepackage{placeins}
\title{Punctuation restoration Model and Spacing model for Korean Ancient Document
\thanks{\textit{\underline{Citation}}: 
\textbf{Authors. Title. Pages.... DOI:000000/11111.}} 
}

\author{
  Taehong Jang \\
  Nara Information \\
  starbirdnara@gmail.com \\
  \And
  Joonmo Ahn \\
  Nara Information \\
  jmahn.nara@gmail.com \\
  \And
  Sojung Lucia Kim \\
  Nara Information \\
  sojung.kim@snu.ac.kr \\
}

\begin{document}
\maketitle

\begin{abstract}
%현재 도입부와 내용 동일함, 수정할 것임
In Korean ancient documents, there is no spacing or punctuation, and they are written in classical Chinese characters. This makes it challenging for modern individuals and translation models to accurately interpret and translate them. While China has models predicting punctuation and spacing, applying them directly to Korean texts is problematic due to data differences. Therefore, we developed the first models which predict punctuation and spacing for Korean historical texts and evaluated their performance. Our punctuation restoration model achieved an F1 score of 0.84, and Spacing model achieved a score of 0.96. It has the advantage of enabling inference on low-performance GPUs with less VRAM while maintaining quite high accuracy.
\end{abstract}

% keywords can be removed

\section{Introduction}
Korean ancient documents are written in Chinese characters, and they do not contain any spacing or punctuation marks. Therefore, it is difficult for modern readers to understand their meaning. Additionally, segmenting the text and incorporating punctuation marks enhances the performance of translation tasks\cite{Park2023translation, Mogahed2012Arabic}. 
However, there is a lack of models in Korea that can create spacing and punctuation for such texts\cite{Park2023translation}.
To address this, we have developed two models by adding additional layers to the BERT model and trained them. One model was trained using data created for predicting punctuation, while the other used data created for predicting spacing. The data were processed from the same source materials, which include various data related to Confucian classics and the Annals of the Joseon Dynasty.

\section{Model}

\subsection{About models}
\begin{figure}[ht]
  \centering
  \begin{minipage}[b]{0.45\linewidth}
    \includegraphics[width=\linewidth]{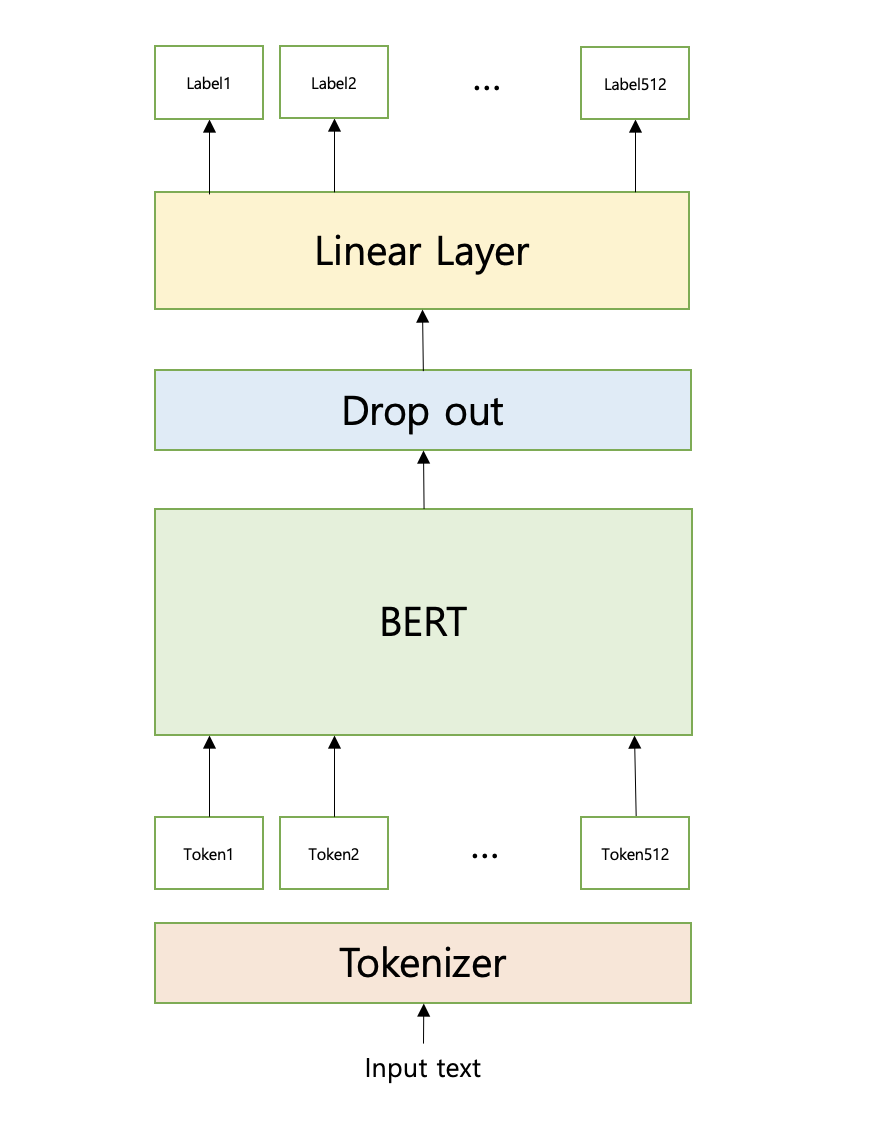}
    \caption{The structures of punctuation/spacing Models}
    \label{fig:fig1}
  \end{minipage}
  \hfill
  \begin{minipage}[b]{0.45\linewidth}
    \includegraphics[width=\linewidth]{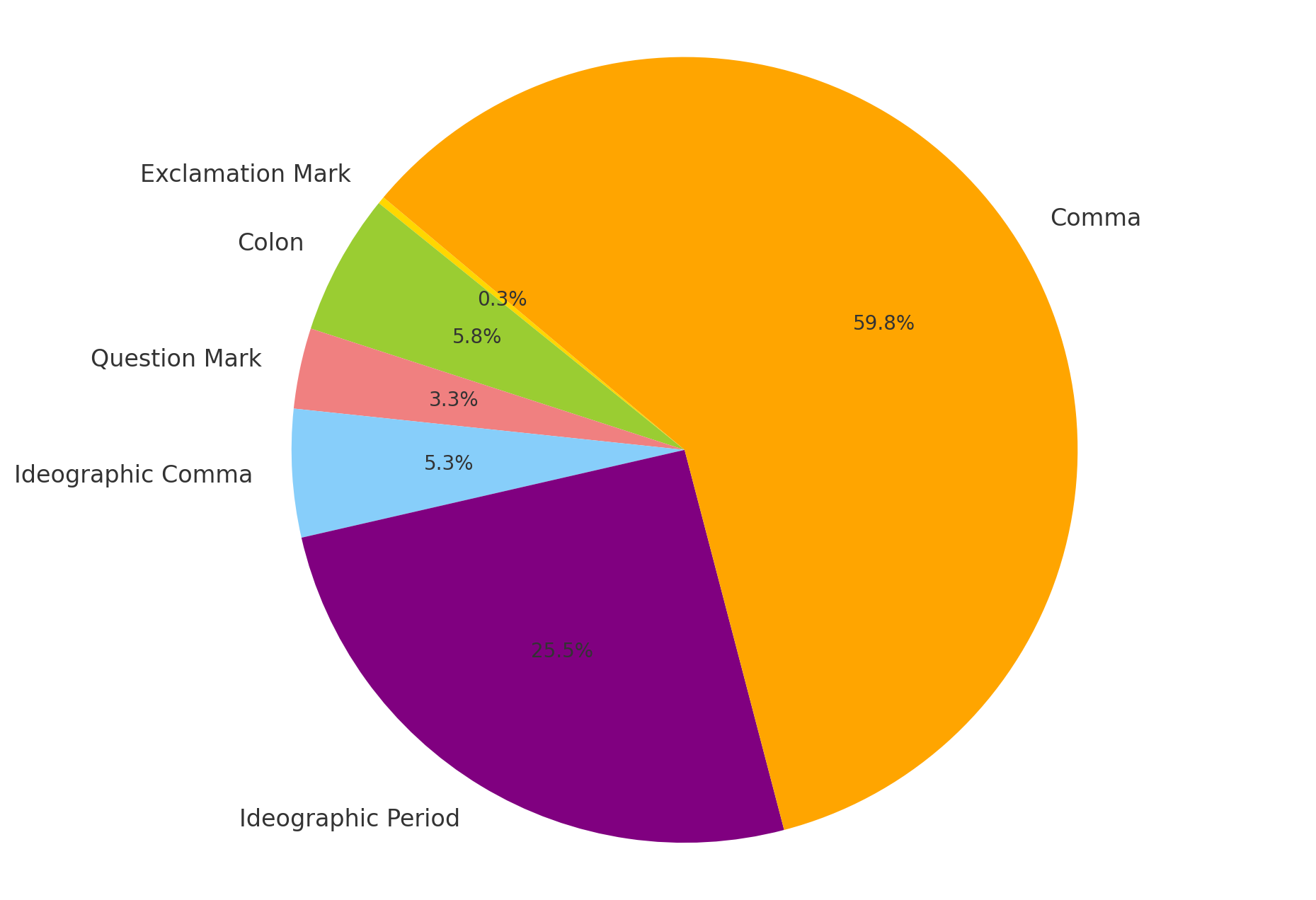}
    \caption{The proportion of punctuation marks}
    \label{fig:fig2}
  \end{minipage}
\end{figure}

We used BERT model\cite{devlin2019bert} as a pre-trained model and created our model by adding a linear layer. The BERT model excels in understanding context because of using the Transformer architecture\cite{vaswani2023transformer}. Additionally, it is easy to fine-tune, making it adaptable to various tasks, and it demonstrates strong performance across multiple natural language processing tasks. We used "bert-base-chinese" model \footnote{https://huggingface.co/bert-base-chinese}, which was pretrained by Chinese. This model consists of 12 transformer layers and has a hidden size of 768. Having 768 hidden sizes means that in our model, each token is represented by 768 size of vector. 'bert-base-chinese' has 110 million parameters. We adapted the model to token classification for predicting punctuation and predicting spacing. Figure \ref{fig:fig1} shows the structure of our model.
One(model A) of the two models is trained to predict punctuation. It predict 7 labels('O'\footnote{Label 'O' mean no punctuation markers.}, '!', ':', '?', 'U+FF0C'\footnote{This is the Chinese comma, used to separate clauses within a sentence, similar to the comma in English.}, 'U+3001'
\footnote{Known as the ideographic comma, this character is used in Japanese and Chinese to list items within a sentence.}and 'U+3002'\footnote{This is the Chinese and Japanese full stop, used to mark the end of a sentence, similar to the period in English.}) The other model(Model B) is trained to predict spacing. It predict only 2 labels spacing('\_') and not spacing('O'). 

\subsection{Tokenizer}
We used the tokenizer of 'bert-base-chinese,' which has a vocabulary size of 21,128. Considering the ideographic nature of Chinese characters, the tokenizer treats each Chinese character as a single token. There are some characters not included to tokenizer, so we added 8,323 tokens more to tokenizer's vocabulary.

\section{Datasets}
We used Annals of the Joseon Dynasty(1392-1863)(AJD) and the data related to Confucianism(DRC) that have been digitized by an organization called the Confucian Compilation Center. We split all the text to ensure that each piece did not exceed 512 tokens, and by combining the two datasets, we got about 665,000 sequences in total. There was a time difference in acquiring the two datasets. Initially, we used only the AJD for training the model and also evaluated the model's performance using only the AJD . After getting the DRC data, we then used both datasets for training and evaluated the performance. In both cases, we utilized 90\% of the data for training and assessed the performance using the remaining 10\%. 

\subsection{Annals of the Joseon Dynasty}
The Annals of the Joseon Dynasty, known as the "Joseon Wangjo Sillok" in Korean, are a collection of detailed records of the Joseon Dynasty, which ruled Korea from 1392 to 1863. These annals document the day-to-day affairs of the court, including politics, diplomacy, military matters, the economy, and cultural events. They are notable for their extensive coverage and historical accuracy, providing valuable insights into the era. Compiled by official historians, these records are an important part of Korean heritage and are recognized as a UNESCO World Heritage for their cultural significance. 
After processing AJD, we got more than 412,000 sequences. 

\subsection{Data related to Confucianism}
According to \cite{an2020confucian}, as of 2020, 145 types of books have been digitized, encompassing a total of 30 million characters. Here, the data related to Confucianism is used in a broad sense, not limited to just 'classics' or 'Confucian studies,' but also encompassing various texts authored by Confucian scholars. This encompasses a wide range of genres, from classics to literary works. Some of the data is in Korean, so we have used sentences composed only of Chinese characters. After processing the data, we obtained over 250,000 sequences. 

\subsection{Preprocessing}

When preprocessing the data, there were two main issues:
\begin{enumerate}
    \item The two datasets were organized by different institutions, leading to slight differences in the way punctuation marks were annotated.
    \item Deciding how to set labels when using a token classification model.
\end{enumerate}

Considering these, we sought expert advice for the preprocessing. Regarding the first issue, we unified or removed inconsistent punctuation marks. For instance, the two datasets used different notations for representing quotations within quotations, brackets denoting book titles, and markings for damaged text. Fortunately, the 6 punctuation marks we ultimately chose as labels were significant and used similarly in both datasets. 

When deciding on labels for punctuation marks, we encountered two problems: (1) when a punctuation mark is located before a token, and (2) when punctuation marks appear consecutively. An example of (1) is a quotation mark that denotes quotation's beginning. Typically, models predict punctuation marks that come after a token, but predicting those that come before leads to a structurally paired format of previous and following labels. In the worst case, the number of labels could be squared, making the data for each label sparse and negatively impacting learning. Additionally, consecutive punctuation marks, while not numerous, can compound with (1) to further increase the number of cases and sparsity. Therefore, after consulting with experts, we processed the data as follows.

More than 20 punctuation marks in the datasets were reduced to 6 punctuation marks by removal and replacement\cite{nagy2021puncbert}. Semicolons are used to separate sentence, so they were replaced to 'U+3002' character(Ideographic period). Quotation marks can appear nested within other quotation marks, and their representation varies depending on the number of nestings, so we unified the quotation marks into a single mark. We divided it into two cases and processed the mark differently for each. If the mark directly followed other punctuation marks, we deleted it. In the other case, we removed the quoting function of the quotation marks, leaving only their function as segmenting the content, and replaced them with 'U+FF0C'(Chinese comma) character. Also, we properly deleted brackets and contents in bracket. Brackets are also could appear ahead of tokens. So it is not proper to predict. In addition, through various processing steps, we ensured that there were no punctuation marks preceding a token or appearing consecutively. 

The spacing dataset was derived from the punctuation dataset by simply replacing all 6 punctuation mark to spacing characters("\_"). Both datasets were processed in accordance with the CoNLL format\cite{kim2003conll}.

\section{Training}

We trained the models by 90\% of two datasets, on the NVIDIA A5000 GPU. We allocated 10\% of the 90\% datasets as the validation dataset, ensuring to save the best model when the loss of the validation dataset is minimized. The batch size was 16, and it was trained for 15 epochs. Each epoch took about 4.5 hours. We used AdamW optimizer\cite{loshchilov2019adamw}, and set learning rate to $5 \times{10^{-5}}$. Loss Function was Cross-Entropy Loss. The validation loss was lowest around epoch 10, and the best model was found between epochs 5 and 12.
% 마무리 어색(데이터셋에 따른 모델 구분으로 얘기를 해야할 필요가 있는지?)

\begin{table}
 \caption{Punctuation model's Accuracy}
  \centering
  \begin{tabular}{llll}
    \toprule
    Punctuation mark (Unicode)           & Precision & Recall & F1-Score \\
    \midrule
    Exclamation Mark (U+0021)  & 0.81      & 0.66   & 0.73     \\ 
    Question Mark (U+003F)     & 0.90      & 0.95   & 0.93     \\
    Ideographic Comma (U+3001) & 0.83      & 0.75   & 0.79     \\
    Colon (U+003A)             & 0.95      & 0.95   & 0.95     \\
    Ideographic Period (U+3002)& 0.83      & 0.83   & 0.83     \\
    Comma (U+FF0C)             & 0.84      & 0.85   & 0.84     \\
    \addlinespace
    Total                      & 0.86      & 0.83   & 0.84     \\
    \bottomrule
  \end{tabular}
  \label{tab:table1}
\end{table}

\begin{table}
 \caption{Spacing model's accuracy}
  \centering
  \begin{tabular}{llll}
    \toprule
    Spacing & Precision & Recall & F1-Score \\
    \midrule
    O      & 0.98      & 0.98   & 0.98     \\
    \_(spacing)     & 0.93      & 0.92   & 0.93     \\
    \addlinespace
    Total  & 0.96      & 0.95   & 0.96     \\
    \bottomrule
  \end{tabular}
  \label{tab:table2}
\end{table}

\section{Evaluation}
We evaluated models using F1 score.The test dataset is 10\% of the integrated dataset and has not been used for training. Punctuation model has 7 labels and spacing model has 2 labels. When evaluating the punctuation model, we removed the 'O' label because it overwhelmingly dominates in proportion compared to the other 6 labels. Including the accuracy when predicting the 'O' label could lead to an overestimation of the model's performance. In the case of the spacing model, the 'O' label appeared roughly five times more frequently than '\_' and both labels were included in the performance evaluation. The best punctuation model achieved a score of 0.84 on the F1 score for 6 labels. The best spacing model achieved a score of 0.96. For detailed F1 scores, precision, and recall scores for each label, refer to Table \ref{tab:table1} and Table \ref{tab:table2}. Generally, labels had more data tended to have higher scores. In the case of '?' and ':', despite having relatively less data, it is assumed that the scores were high because specific characters often appeared around these marks. For '?', characters such as \begin{CJK}{UTF8}{mj}如, 何, 孰, 誰 \end{CJK} often appeared together, and for ':', characters like \begin{CJK}{UTF8}{mj}曰\end{CJK} also appeared.

\section{Conclusion}

There have been attempts to translate ancient documents in Korea, and it is said that the presence or absence of punctuation and segmentation greatly affects the performance of the translation\cite{Park2023translation}. In China, there are models for segmentation and punctuation\cite{han2018segment}, but the punctuation marks and the expression of proper nouns differ from those in Korea. Therefore, it is necessary to train using Korean ancient document data. We trained the first punctuation and segmentation models for Korean classics in Korea. The punctuation model showed an accuracy of 0.84 in terms of the F1 score, while the segmentation model demonstrated an accuracy of 0.96. Our models will be helpful in translating ancient documents that lack punctuation and segmentation. Recently, with the emergence of GPTs, implementing models specialized in NLP has become more feasible. We requested a model specialized in ancient document punctuation from GPTs, and tried to predict punctuation by removing the punctuation marks from sentences that were already punctuated. When compared with the original ground truth, it showed an F1 score of approximately 0.84, demonstrating a performance similar to that of our model.Our model shows a performance similar to that of GPTs while using only 1.5GB of VRAM. It's said that training the smallest 7B model of LLaMA2\cite{touvron2023llama2} requires around 28GB of VRAM\cite{chen2023netgpt}, while inference would need less resources. GPTs, based on GPT-4, don't have disclosed VRAM usage for inference\cite{openai2023gpt4}, but it's presumed to use more resources than the LLaMA 7B model. Thus, our model can be considered efficient in its GPU usage. Additionally, we are considering various methods to improve our model's performance. This includes strategies for addressing label imbalance\cite{megahed2021imbalance} and the collection and integration of additional ancient document data. We will attempt to develop a translation model that performs well even without punctuation processing.

\section*{Acknowledgments}
This was supported in part by Young-ho,Son and Bong-nyu,Kwon

%Bibliography
\bibliographystyle{unsrt}  
\bibliography{references}

\end{document}